\documentclass{article}

    \PassOptionsToPackage{numbers}{natbib}

    \usepackage[final]{neurips_2024}

\usepackage[utf8]{inputenc} 
\usepackage[T1]{fontenc}    
\usepackage{hyperref}       
\usepackage{url}            
\usepackage{booktabs}       
\usepackage{amsfonts}       
\usepackage{nicefrac}       
\usepackage{microtype}      
\usepackage[table]{xcolor}         
\usepackage{graphicx}
\usepackage{amsmath}
\usepackage{mathrsfs}
\usepackage{multirow}
\usepackage{caption}

\usepackage{amssymb}  
\usepackage{pifont}   

\newcommand{\cmark}{\ding{51}}
\newcommand{\xmark}{\ding{55}}

\def\modelname{\texttt{EchoSegnet}}
\def\aisrmname{\texttt{AISRM}}
\def\benchname{\textbf{3DAVS-S34-O7}}
\def\benchnamenorm{{3DAVS-S34-O7}}

\title{3D Audio-Visual Segmentation}

\author{%
  Artem Sokolov \quad Swapnil Bhosale \quad Xiatian Zhu \\
  University of Surrey, UK \\
  \texttt{\{as05633, s.bhosale, xiatian.zhu\}@surrey.ac.uk} \\
}

\begin{document}

\maketitle

\begin{abstract}
  Recognizing the sounding objects in scenes is a longstanding objective in embodied AI, with diverse applications in robotics and AR/VR/MR. To that end, Audio-Visual Segmentation (AVS), taking as condition an audio signal to identify the masks of the target sounding objects in an input image with synchronous camera and microphone sensors, has been recently advanced. However, this paradigm is still insufficient for real-world operation, as the mapping from 2D images to 3D scenes is missing. To address this fundamental limitation, we introduce a novel research problem, {\em 3D Audio-Visual Segmentation}, extending the existing AVS to the 3D output space. This problem poses more challenges due to variations in camera extrinsics, audio scattering, occlusions, and diverse acoustics across sounding object categories. To facilitate this research, we create the very first simulation based benchmark, {\em 3DAVS-S34-O7}, providing photorealistic 3D scene environments with grounded spatial audio under \textit{single-instance} and \textit{multi-instance} settings, across 34 scenes and 7 object categories. This is made possible by re-purposing the Habitat simulator \cite{habitat19iccv} to generate comprehensive annotations of sounding object locations and corresponding 3D masks. Subsequently, we propose a new approach, \modelname{}, characterized by integrating the ready-to-use knowledge from pretrained 2D audio-visual foundation models synergistically with 3D visual scene representation through spatial audio-aware mask alignment and refinement. Extensive experiments demonstrate that \modelname{} can effectively segment sounding objects in 3D space on our new benchmark, representing a significant advancement in the field of embodied AI. Project page: \href{https://x-up-lab.github.io/research/3d-audio-visual-segmentation/}{https://x-up-lab.github.io/research/3d-audio-visual-segmentation/}
\end{abstract}

\section{Introduction}
 
Human perception of the real world, both visual and acoustic, predominantly occurs in three dimensions. Prior psychology literature \cite{welch1980immediate} has highlighted humans' remarkable ability to correspond across multiple modalities, often involving the association of events across these modalities. For instance, we can effortlessly ground emergent surround sound with its potential source in 3D visuals \cite{mo2024weakly}.
Inspired by this capability, a crucial aspect in the development of embodied AI systems is their ability to integrate cues from synchronous multimodal input streams and establish targets corresponding to their goals. In this work, we aim to build a machine model to achieve this multimodal correspondence, particularly targeted towards the task of audio-visual segmentation (AVS) in 3D.

\begin{figure}[t]
    \centering
    \includegraphics[width=1.0\textwidth]{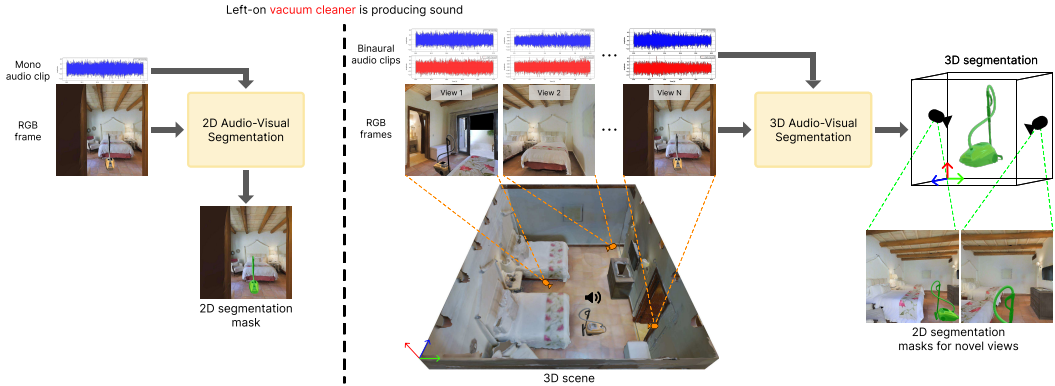}
    \caption{
    Comparison of the existing 2D AVS task with our proposed 3D AVS.
    Former task utilizes single channel audio to generate pixel-level masks of the potential sounding object in the input RGB frame.
    3D AVS on the other hand is aimed at generating 3D masks (from which multi-view consistent 2D masks can be rendered) while utilizing multichannel (spatial) audio.
    }
    \label{fig:3davseg}
\end{figure}

Albeit AVS has been widely explored within audio-visual scene analysis and correspondence learning, prominent research in this field has focused on 2D environments involving mono (single channel) sound sources, thus devoid of spatial presence entirely.
In this paper, we take the first step towards exploring 3D AVS and introduce a large benchmark, \benchname{}. Our exploration is rooted in a fundamental grounding problem: given an embodied agent equipped with a camera and a binaural microphone, can we teach the agent to obtain fine-grained localization of potential sounding objects (generally by predicting a segment-level mask of the object in 3D) while also utilizing spatial audio cues? (see Fig. \ref{fig:3davseg})
Furthermore, we extend our benchmark to include a more competitive \textit{multi-instance} setup where, although multiple instances of the same object might be present in the scene, the goal is to segment only the sounding instance. This setup helps us testify to the efficacy of spatial presence harnessed from the input binaural audio samples.

Recently, 3D Gaussian Splatting (3D-GS) \cite{3dgs} has emerged as a prospective method for modeling static 3D scenes directly from input RGB frames. Owing to its explicit Gaussian based representation, it has paved a natural pathway for 3D visual segmentation \cite{hu2024sagdboundaryenhancedsegment3d, ye2023gaussian, shen2024flashsplat}.
Deriving inspiration from human spatial memory in indoor environments, we design \modelname{}, a purely training-free pipeline for 3D AVS within a 3D-GS representation. \modelname{} leverages 2D foundation models (namely SAM \cite{kirillov2023segany} and ImageBind \cite{girdhar2023imagebind}) to first obtain 2D AVS masks on the input RGB frames.
These 2D AVS masks are further used to segment the Gaussians in the learned 3D-GS representation to obtain multi-view masks to achieve a consistent 3D segmentation.

To summarize, we make the following \textit{contributions}: 
(1) the first 3D audio-visual segmentation benchmark composing of fairly complex indoor room scenes with integrated spatial sound cues;
(2) a training-free AVS framework, \modelname{}, capable of syncing across sequential frames from 3D environments;
(3) a novel Audio-Informed Spatial Refinement Module \aisrmname{}, designed to enhance 3D segmentation and resolve ambiguities in complex, multi-instance environments by leveraging spatial audio intensity maps.
We perform a comprehensive evaluation of \modelname{} on the proposed \benchname{} for both \textit{single-instance} and \textit{multi-instance} scenarios, along with an ablative comparison with existing 2D AVS models, highlighting their shortcomings in aligning audio-visual cues within 3D scenes -establishing their adaptation to \benchname{} as non-trivial.

\section{Related Work}
 
\noindent\textbf{Audio-visual segmentation}
Existing AVS methods cater to 2D scenes with mono audio as inputs to identify audible visual pixels associated with a given audio signal \cite{zhou2022avs, hao2023improvingaudiovisualsegmentationbidirectional, mao2023contrastiveconditionallatentdiffusion, mo2023avsamsegmentmodelmeets, shi2024crossmodalcognitiveconsensusguided, Zhou2022ContrastivePS} and are typically trained on thousands of manually annotated 2D segmentation masks. Although there have been recent improvements on reducing the dependence on annotated AV masks using weakly supervised \cite{mo2024weakly, liu2024annotation} or entirely unsupervised methods \cite{chen2024unraveling, liu2023bavsbootstrappingaudiovisualsegmentation, yang2024cooperation, fang2024audio}, there has not been any effort extending the task of AVS particularly to 3D scenes, with spatial audio cues. 
To address this gap, we propose the first benchmark for 3D AVS harnessing existing embodied AI platforms (Habitat simulator \cite{habitat19iccv}) to capture visual and (binaural) acoustic cues for sounding objects placed in 3D indoor scenes. We believe this lays a prominent groundwork for systematic evaluation of future embodied systems for 3D segmentation.

\noindent\textbf{3D scene representations}
Point-based rendering techniques, initiated by \cite{grossman1998point}, utilize point-based explicit representation where each point affects a single pixel. Zwicker et al. \cite{zwicker2001surface} advanced this with ellipsoid-based rendering (splatting), allowing mutual overlap to fill image holes.
In the absence of given geometry, Mildenhall et al. \cite{nerf} explored neural implicit representation, NeRF, predicting view-dependent radiance via implicit density fields.
3D Gaussian Splatting (3D-GS) \cite{3dgs}, a novel-view synthesis method, employs explicit point-based representation, contrasting with NeRF's volumetric rendering. 
Owing to its real-time high-quality rendering capabilities, 3D-GS has been applied to various domains, including simultaneous localization \cite{keetha2023splatam, matsuki2023gaussian}, content generation \cite{tang2023dreamgaussian}, and 4D dynamic scenes \cite{li2023spacetime, wu20234d}, among others.
In this work, we utilize the explicit representation from learned 3D-GS and decompose 2D masks of potential sounding objects to obtain consistent 3D masks.

\section{Dataset}
 
Our proposed \benchnamenorm{} is profoundly motivated towards simulating real-world indoor scenes, in terms of the visual quality of the scenes as well as the acoustic response generated by the objects placed within it.
In the context of our 3D AVS task, we define an observation as $O$=$\{(v_i, a_i, m_i)\}_{i=1}^{n}$, where $v_i$, $a_i$ represent the visual (RGB view, $\mathbb{R}^{1008 \times 1008 \times 3}$) and acoustic (1 second binaural audio, at 44.1kHz) cues respectively, captured by the embodied agent at $i$-th time. $m_i$ represents a binary mask corresponding to $v_i$ highlighting the sounding object.
To record an observation, we load a randomly sampled scene from the Habitat-Matterport3D dataset \cite{ramakrishnan2021habitatmatterport3ddatasethm3d} into the SoundSpaces 2.0 \cite{chen22soundspaces2}. Next, we place a semantically relevant sounding object (for instance, \textit{bathroom}$\leftrightarrow$\textit{washing machine}, \textit{kitchen}$\leftrightarrow$\textit{microwave}, etc.) which emits a sound based on a mono audio (corresponding to the placed object and sourced from \cite{font2013freesound, bbc_sound_effects, piczak2015dataset}).
We capture $n = 120$ frames at 1 fps symbolizing different positions along the moving agent's path. 
\begin{figure}
  \noindent
  \begin{minipage}[t]{.35\linewidth} 
    \captionof{figure}{Sample scenes from \benchname{} dataset.} 
    \includegraphics[width=0.95\linewidth]{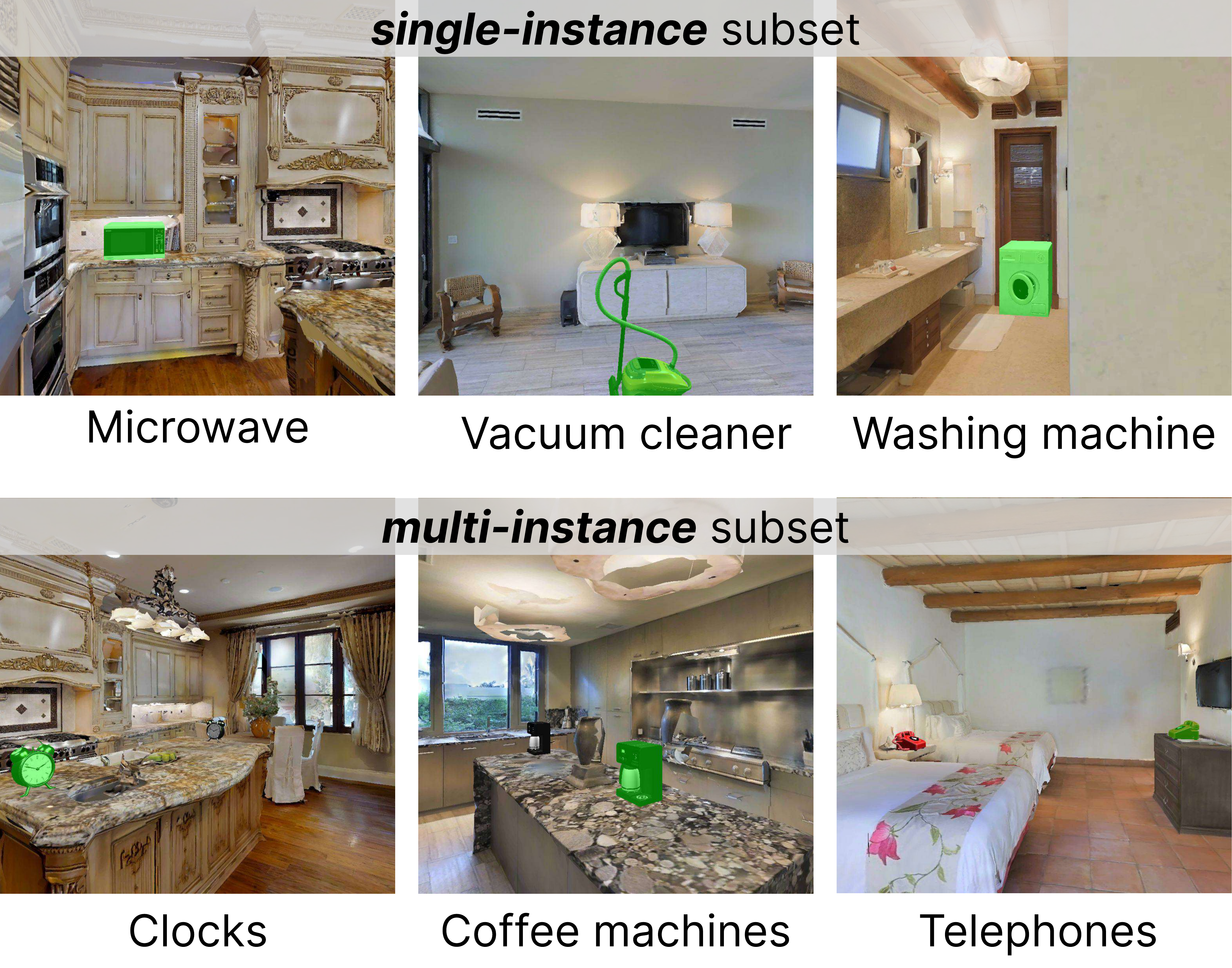} 
    \label{fig:sample_scenes}
  \end{minipage}
  \hspace{0.02\linewidth} 
  \begin{minipage}[t]{.60\linewidth} 
    \centering
    \captionof{table}{Comparison with existing 3D visual segmentation benchmarks: NVOS \protect{\cite{ren-cvpr2022-nvos}} and SPIn-NeRF \protect{\cite{spinnerf}} support promptable 3D visual segmentation but lack spatial audio.}
    \label{tab:benchmark_comparison}
    \begin{tabular}{@{}l|c|c|c@{}}
      \toprule
      Benchmark & \#(Objects) & \#(Scenes) & Audio \\ \midrule
      NVOS \cite{ren-cvpr2022-nvos} & 8 & 8 & \xmark \\
      SPIn-NeRF \cite{spinnerf} & 6 & 10 & \xmark \\ 
      \benchname{} (Ours) & 7 & 34 & \cmark \\ \midrule
      \multicolumn{1}{@{}l}{\hspace{1em}\textit{single-instance}} & 7 & 25 & \cmark \\
      \multicolumn{1}{@{}l}{\hspace{1em}\textit{multi-instance}} & 7 & 9 & \cmark \\
      \bottomrule
    \end{tabular}
  \end{minipage}
\end{figure}
Alongside the above \textit{single-instance} setup, we also explore a slightly challenging \textit{multi-instance} setup wherein, we place multiple instances of the sounding object, although only one instance is sound-emitting (Fig. \ref{fig:sample_scenes}).
We split each observation into 7:1 for train:test split (following \cite{barron2022mipnerf360}). Details of the selected scanned spaces and sound-emitting objects are provided in Appendix \ref{sec:dataset_materials_apendix}.
\section{Method}
 
\begin{figure}[t]
    \centering
    \includegraphics[width=0.9\textwidth]{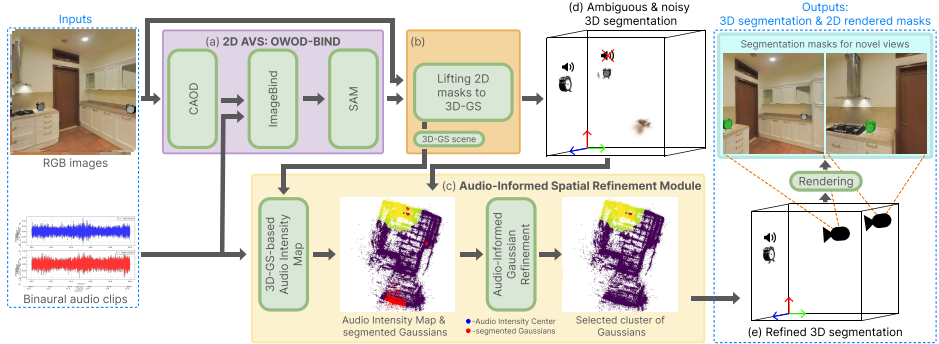}
    \caption{
    Overview of \modelname{}: (a) 2D AVS pipeline OWOD-BIND \protect{\cite{Bhosale2023LeveragingFM}} generates 2D masks. (b) These masks are lifted into a 3D-GS scene representation using \protect{\cite{hu2024sagdboundaryenhancedsegment3d}} with a modified voting strategy. (d) The initial 3D segmentation may contain noise and ambiguities, as spatial relationships between objects and sound were not considered. (c) To address this, we apply the novel Audio-Informed Spatial Refinement Module (\aisrmname{}). (e) In the refined 3D segmentation, only the sound-emitting object instance is retained, and noise is filtered out.
    }
    \label{fig:echosegnet}
\end{figure}

Considering the complexity of the 3D AVS task and deriving inspiration from human spatial memory, we propose \modelname{} (see Fig. \ref{fig:echosegnet}), a training-free pipeline leveraging 2D foundation models.
For each input view, $v_i$, we first obtain corresponding 2D AVS masks $\hat{m}_i$ using OWOD-BIND \cite{Bhosale2023LeveragingFM}. 
Particularly, OWOD-BIND prompts the SAM \cite{kirillov2023segany} model using bounding boxes obtained from class-agnostic object detection (CAOD) \cite{Maaz2022Multimodal}. The mask proposals from SAM are further filtered based on maximum cosine similarity with $a_i$'s audio embedding generated using ImageBind \cite{girdhar2023imagebind}.

Please note, the masks $\hat{m}_i$ are confined to $v_i$ however
the main goal of the 3D AVS task is to obtain multi-view consistent masks of the potential sounding object for novel viewing positions (beyond $v_i$). Moreover, the sounding object may be fully or partly visible in the novel view.
To achieve this,
we propose to lift the 2D AVS masks $\hat{m}_i$ within an explicit 3D scene representation $\mathcal{G}$, generated using vanilla 3D-GS \cite{kerbl3Dgaussians}.
Although similar approaches exist for salient 3D visual segmentation (such as \cite{hu2024sagdboundaryenhancedsegment3d}), the sounding object in the context of 3D AVS, may not always be salient (i.e in the foreground). As a result, unlike \cite{hu2024sagdboundaryenhancedsegment3d}, we opt to exclude out-of-view projections from the voting process for selecting underlying Gaussians as well as directly lift $\hat{m}_i$ (see Appendix \ref{sec:modified_sagd_lifting_appendix}).

\noindent\textbf{Audio-Informed Spatial Refinement Module (\aisrmname{})}
Although the above lifting process yields 3D segmentation masks, we observe certain ambiguities: (1) in the case of a \textit{multiple-instance} setup, computation of $\hat{m}_i$, being devoid of spatial audio, is unable to accurately localize only the sound-emitting instance of the object (see Fig. ~\ref{fig:echosegnet}(d)), and (2) due to errors in audio-visual alignment within the frozen ImageBind, $\hat{m}_i$ often includes other (silent) objects in the vicinity of the sound-emitting object.

To handle both the ambiguities, we start with a \textbf{3D-GS-based Audio Intensity Map}. Specifically, we introduce additional labels \( I_{\mathbf{g}}\) on every Gaussian $\mathbf{g}$ within our scene representation $\mathcal{G}$ by weighing the root mean square (RMS) intensities on the agent's left and right audio channels, $R^l$, $R^r$ respectively.
For each Gaussian $\mathbf{g}$, we compute $I_{\mathbf{g}} = \sum_{i=1}^{t} \frac{|R_i^l - R_i^r|}{\max(R_i^l, R_i^r)} \cdot \mathbb{I}_{\text{RMS}}(\mathbf{g}_{center}, a_i)$, where \( \mathbb{I}_{\text{RMS}}(.) \) equals 1 if the Gaussian center $\mathbf{g}_{center}$ is located on the side with the greater RMS intensity based on the binaural audio observation \( a_i \).
\cite{yang2024rila} proposed a similar intensity map but in two dimensions, and not grounded within an underlying 3D-GS representation.

We then perform an \textbf{Audio-Informed Gaussian Refinement} process through spatial clustering, guided by $I_{\mathbf{g}}$. We cluster the segmented 3D Gaussians using DBSCAN \cite{10.5555/3001460.3001507} and filter clusters with volumes $> \mu_v + 0.5 \sigma_v$ where \( \mu_v \) is the mean volume and \( \sigma_v \) is the standard deviation of all cluster volumes.
Next we localize the audio intensity center by computing an average of the Gaussian center coordinates weighted by $I_\mathbf{g}$ (only Gaussians with $I_\mathbf{g} > \tau_{\text{ref}}$ are considered).
We hypothesize that the cluster closest to the computed audio intensity center consists of Gaussians corresponding to the sound-emitting object, effectively filtering out both the inclusion of silent objects, as well as non-sound-emitting instances in the multi-instance setting.

\section{Experiments}
\label{sec:experiments}
 
In this section, we demonstrate the effectiveness of \modelname{} on the \benchname{} benchmark, with a particular focus on the contribution of \aisrmname{}.
Following \cite{zhou2022avs}, we adopt mIoU and F-Score as the metrics to estimate the segmentation performance. For implementation details, please refer to Appendix \ref{sec:implementation_details_appendix}. 
\begin{figure}[t]
    \centering
    \includegraphics[width=1.0\textwidth]{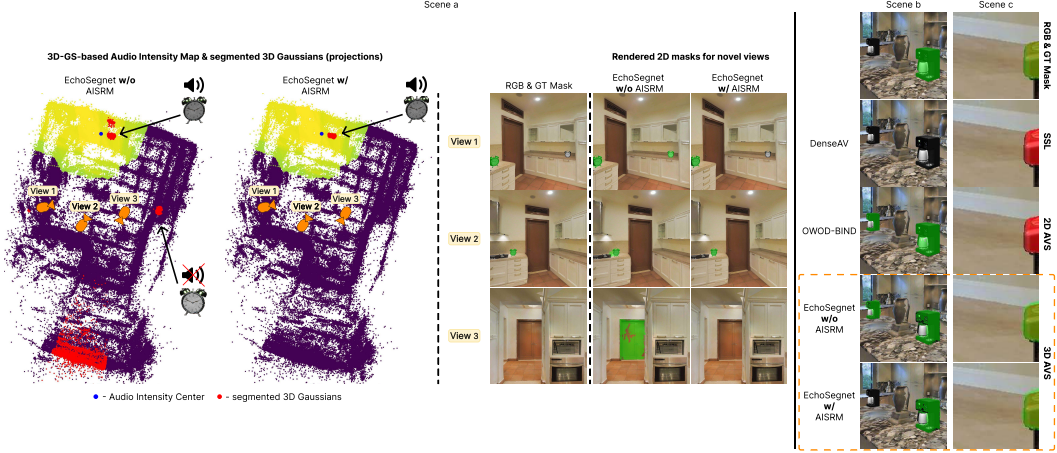}
    \caption{
    Left: (Scene a) Qualitative comparison of \modelname{} performance with and without \aisrmname{}, illustrated through projected 3D-GS scene representation and renderings. Right: Comparison between DenseAV (SSL), OWOD-BIND (2D AVS) and  \modelname{}. (Scene b) OWOD-BIND \protect{\cite{Bhosale2023LeveragingFM}} incorrectly segments the non-sound-emitting coffee machine. (Scene c) Both SSL and 2D AVS fail to handle a complex scenario where only a small part of the sound-emitting telephone is present in the view, whereas \modelname{} successfully addresses this challenge. 
    }
    \label{fig:qualitative_results}
\end{figure}
\begin{table}[h]
  \centering
  \caption{Performance comparison of \modelname{} (3D AVS) with and without \aisrmname{}, and comparison against 2D AVS and SSL pipelines on both subsets of the \benchname{} benchmark.}
  \label{tab:combined_results}
  \begin{tabular}{@{}l|cc|cc@{}}
    \toprule
    Approach & \multicolumn{2}{c|}{\textit{single-instance}} & \multicolumn{2}{c}{\textit{multi-instance}} \\ 
    \cmidrule(lr){2-3} \cmidrule(lr){4-5}
                      & mIoU \(\uparrow\) & F-Score \(\uparrow\) & mIoU \(\uparrow\) & F-Score \(\uparrow\) \\ \midrule
    \modelname{} w/o \aisrmname{}  & 0.761 & 0.628 & 0.757 & 0.609 \\
    \modelname{} w/ \aisrmname{}   & \textbf{0.823} & \textbf{0.730} & \textbf{0.801} & \textbf{0.714} \\ \midrule
    DenseAV \cite{hamilton2024separating} (2D SSL)    & 0.426 & 0.023 & 0.436 & 0.023 \\
    OWOD-BIND \cite{Bhosale2023LeveragingFM} (2D AVS) & 0.693 & 0.523 & 0.696 & 0.502 \\
    \bottomrule
  \end{tabular}
\end{table}
From Table \ref{tab:combined_results}, it is evident that removing the \aisrmname{} module results in a performance drop across both \textit{single-instance} and \textit{multiple-instance} settings.
For the \textit{single-instance} setting, mIoU decreases by 0.06, and F-Score drops by 0.10. Similarly, in the \textit{multiple-instance} setting, mIoU drops by 0.04, and F-Score decreases by 0.11. As illustrated in Figure \ref{fig:qualitative_results} (Left), omitting \aisrmname{} introduces noisy Gaussians representing silent objects (e.g., \textit{the door}, view 3), negatively impacting performance across both subsets. In the \textit{multiple-instance} setting, the inability to distinguish between sound-emitting and non-sound-emitting instances of the same object further reduces segmentation accuracy (e.g., both \textit{clocks} are segmented, but only one is sound-emitting, view 1). Additionally, Gaussians in the vicinity of the sounding object (clock) are also incorrectly segmented (view 2). 

\noindent \textbf{Comparison Between SSL, 2D, and 3D Audio-Visual Segmentation.} 
We propose \modelname{} as the first approach towards the novel 3D AVS task. Naturally, comparing the performance of existing 2D AVS (and Sound Source Localization (SSL)) approaches for the 3D AVS task is essential to establish the efficacy of \modelname{}. From Table \ref{tab:combined_results}, it can be clearly observed that \modelname{} consistently outperforms 
OWOD-BIND (a 2D AVS method) across both subsets, while 
DenseAV (a SSL method) 
shows significantly poorer and incomparable performance. The strength of \modelname{} in performing 3D AVS lies in its ability to capture spatial relationships between objects and their sounds, which the existing 2D AVS methods lack, often resulting in segmentation of all visible instances (Figure \ref{fig:qualitative_results}, Right, Scene b). 

\section{Conclusion}
 
In this work, we introduced 3D Audio-Visual segmentation (3D AVS) as a novel extension of the existing 2D AVS paradigm. 
We presented the \benchname{} benchmark, the first simulation-based large dataset for 3D AVS, featuring photorealistic environments with spatial audio across 34 scenes and 7 object categories. 
Our proposed method, \modelname{}, effectively segments sounding objects in 3D scenes in a training-free pipeline leveraging 2D audio-visual foundation models and 3D Gaussian Splatting. 
We believe this marks a significant advancement in bridging the gap between 2D and 3D audio-visual understanding, with broader implications for embodied AI.
Looking ahead, we aim to explore diverse acoustic environments and dynamic objects as the future scope of this work.

\newpage
\bibliographystyle{plainnat}
\bibliography{references}

\newpage  

\appendix

\section{Appendix}

\subsection{Limitations \& Failure Cases}
\label{sec:limitations_failure_cases_appendix}

\begin{figure}[ht]
\centering
\includegraphics[width=1.0\textwidth]{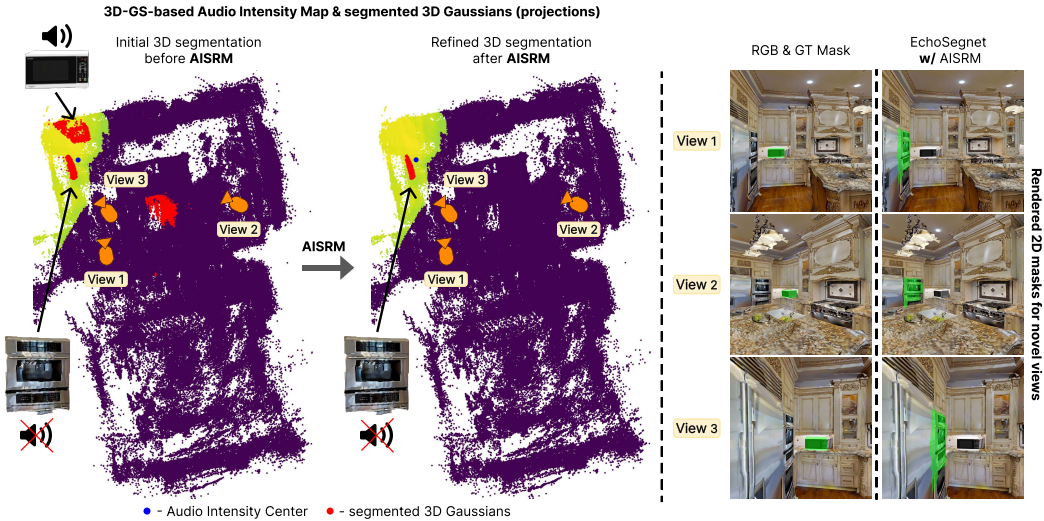}
\caption{
Failure case: Due to the close proximity of the microwave and oven, the \aisrmname{} mistakenly refines the segmented Gaussians to those of the silent oven, discarding the Gaussians of the sound-emitting microwave.
}
\label{fig:failure_case_1}
\end{figure}

Despite the evident improvements brought by \aisrmname{}, it struggles when objects are positioned too closely, which consequently impacts the overall performance of \modelname{}. In Figure \ref{fig:failure_case_1}, both the microwave and oven are initially segmented in 3D due to a misalignment in ImageBind \cite{girdhar2023imagebind}, even though only the microwave is emitting sound. While \aisrmname{} typically resolves such ambiguities, in this case, the 3D-GS-based Audio Intensity Map provided conflicting guidance due to the proximity of the objects. Consequently, the \aisrmname{} refinement process incorrectly retained the Gaussians corresponding to the silent oven, rather than the sound-emitting microwave, as seen in the rendered 2D masks for novel views.

\subsection{Acknowledgment}
\label{sec:acknowledgement_appendix}

The 3D models used in this research were sourced from Sketchfab~\cite{sketchfab} and are available under various open licenses, including Creative Commons, which permit their use in academic research.

\subsection{Dataset: Sound-Emitting Objects and Scanned Spaces}
\label{sec:dataset_materials_apendix}

\begin{figure}[ht]
\centering
\includegraphics[width=1.0\textwidth]{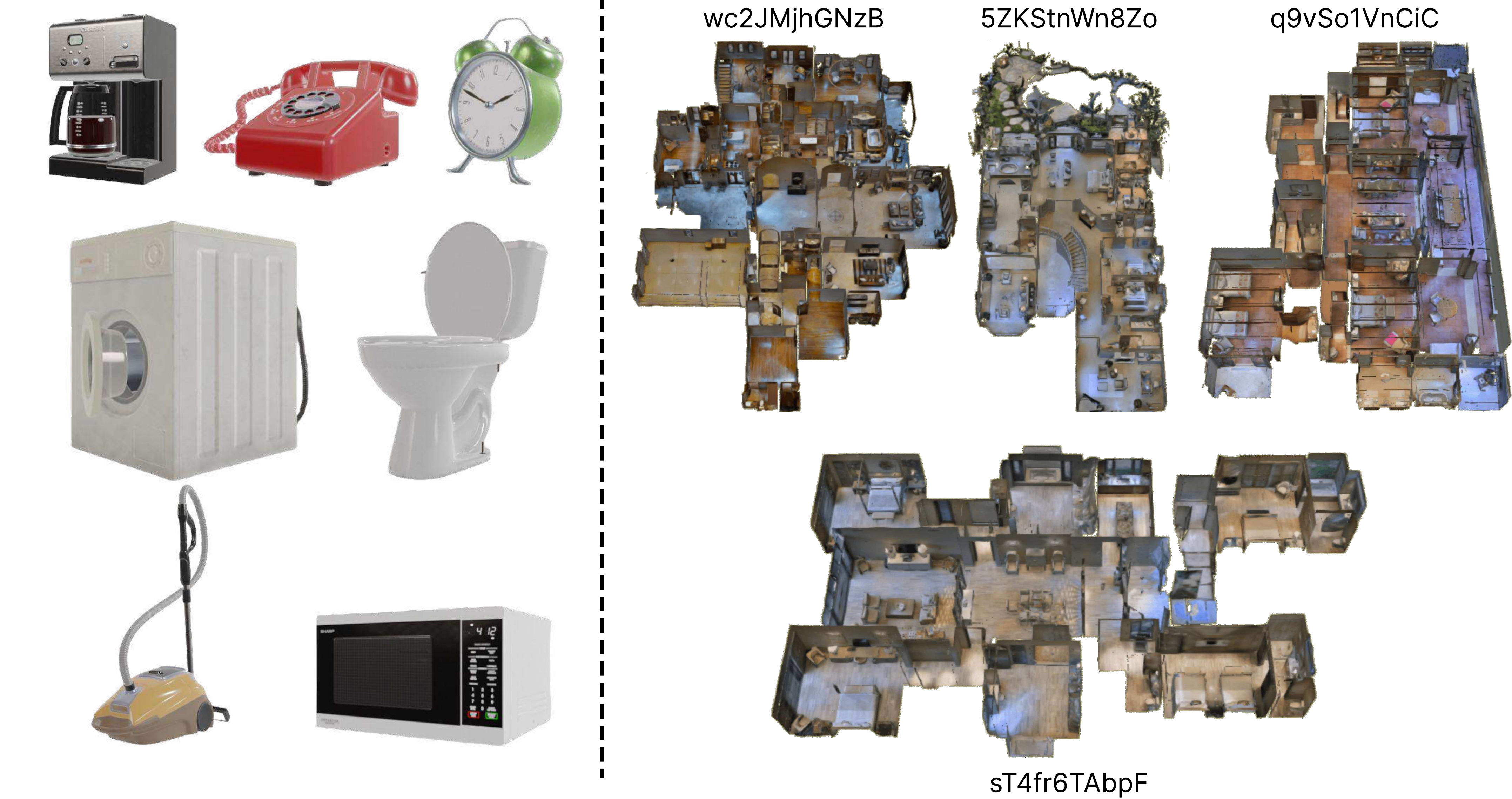}
\caption{
Left: 3D models of the seven selected sound-emitting objects (scale is not preserved). Right: Four selected scanned spaces from Habitat-Matterport3D \protect{\cite{ramakrishnan2021habitatmatterport3ddatasethm3d}} with corresponding dataset IDs.
}
\label{fig:object_spaces}
\end{figure}

The \benchname{} dataset focuses on indoor environments, using scanned spaces from the Habitat-Matterport3D dataset \cite{ramakrishnan2021habitatmatterport3ddatasethm3d}. Four scans were chosen based on their scanning quality and suitability for audio rendering (shown in Fig. \ref{fig:object_spaces}, Right). Seven commonly found sound-emitting objects were selected: a washing machine, toilet, vacuum cleaner, microwave, coffee machine, clock, and telephone (shown in Fig. \ref{fig:object_spaces}, Left). The 3D models of these objects were sourced from Sketchfab \cite{sketchfab} and selected for their realism.

\subsection{Implementation Details}
\label{sec:implementation_details_appendix}

In the OWOD-BIND \cite{Bhosale2023LeveragingFM} pipeline, each 1-second audio clip is extended to 2 seconds by appending 0.5 seconds of audio from neighboring clips before being input into the ImageBind \cite{girdhar2023imagebind} audio encoder, and the threshold \( \tau_{\text{BIND}} \) is set to 0.2. To construct the 3D Gaussians Splatting \cite{kerbl3Dgaussians} scene representation, the original image resolution of 1008x1008 is retained, and each scene is trained for 30,000 iterations. For the modified voting strategy in SAGD \cite{hu2024sagdboundaryenhancedsegment3d}, the threshold \( \tau_{\text{voting}} \) is set to 0.3, with the interval parameter for Gaussian Decomposition fixed at 4, as recommended by \cite{hu2024sagdboundaryenhancedsegment3d}. During the DBSCAN \cite{sander1998density} clustering process, an epsilon value of 0.04 is used, with a minimum point count of 6, as suggested by \textit{Sander et al.} \cite{sander1998density}. For the 3D-GS-based Audio Intensity Map, we use a threshold \( \tau_{\text{ref}} \) of 0.85, meaning only Gaussians with a normalized audio intensity greater than 0.85 are considered. All experiments were conducted using a GeForce GTX 1080 Ti GPU.

\subsection{Modified Voting Strategy for SAGD}
\label{sec:modified_sagd_lifting_appendix}

In contrast to SAGD’s original voting strategy \cite{hu2024sagdboundaryenhancedsegment3d}, which selects 3D Gaussians based on their projection into the 2D object mask more frequently than into the background or out of view, followed by thresholding, we exclude out-of-view projections from the voting process. Thresholding is applied solely based on the ratio of projections into the mask versus the background. This modification allows us to lift object masks as long as the object is consistently segmented in 2D, even if it appears in only a limited number of views. Since we apply additional refinement via \aisrmname{}, compared to the original SAGD \cite{hu2024sagdboundaryenhancedsegment3d}, it is reasonable to use a lower thresholding value \( \tau_{\text{voting}} \). This approach prioritizes segmenting as many Gaussians as possible of the sound-emitting object, even if some Gaussians representing other objects are included, rather than risking the omission of Gaussians related to the sound-emitting object (demonstrated in Table \ref{tab:voting_threshold}). Additionally, we omit SAGD’s original 3D prompt construction strategy, opting instead to directly lift the masks predicted by OWOD-BIND \cite{Bhosale2023LeveragingFM}. 

\begin{table}[h]
  \centering
    \caption{\modelname{} performance on a sample \textit{single-instance} scene (sT4fr6TAbpF, bathroom with sound-emitting vacuum cleaner) with varying \( \tau_{\text{voting}} \) thresholds. Values below 0.3 have little impact on accuracy due to \aisrmname{}, while higher values reduce performance.}
  \label{tab:voting_threshold}
  \begin{tabular}{@{}l|cc@{}}
    \toprule
    \( \tau_{\text{voting}} \) & mIoU \(\uparrow\) & F-Score \(\uparrow\) \\ \midrule
    0.9  & 0.333 & 0.028 \\
    0.8  & 0.333 & 0.028 \\
    0.7  & 0.335 & 0.028 \\
    0.6  & 0.352 & 0.104 \\
    0.5  & 0.396 & 0.241 \\
    0.4  & 0.897 & 0.948 \\
    0.3  & \textbf{0.901} & \textbf{0.949} \\
    0.2  & 0.901 & 0.949 \\
    0.1  & 0.901 & 0.948 \\ \bottomrule
  \end{tabular}
\end{table}

\end{document}